\documentclass[runningheads,orivec]{llncs}
\usepackage[T1]{fontenc}
\usepackage{graphicx}
\usepackage{amsfonts}
\usepackage{amsmath}
\usepackage{xcolor}
\usepackage{multirow}
\usepackage{booktabs}

\makeatletter
\def\ps@headings{\let\@mkboth\@gobbletwo
  \let\@oddfoot\@empty\let\@evenfoot\@empty
  \def\@evenhead{\normalfont\small\hspace{\headlineindent}%
                 \leftmark\hfil}
  \def\@oddhead{\normalfont\small\hfil\rightmark\hspace{\headlineindent}}
  \def\chaptermark##1{}%
  \def\sectionmark##1{}%
  \def\subsectionmark##1{}}
\makeatother
\begin{document}
\title{HERMES: Contrast-Aware Knowledge Graph Reasoning from Clinical Notes for Patient Outcome Prediction}
\titlerunning{HERMES}
%
\author{Gia-Bach Nguyen\inst{1,2} \and Hoang-Ha Nguyen\inst{1} \and
Tuan-Cuong Vuong\inst{3} \and \\ Trang Mai Xuan\inst{3}\thanks{Corresponding author.} \and Duy Quoc Ngo\inst{4} \and Tien-Cuong Nguyen\inst{5} \and \\ Huan Vu\inst{1} \and
Thien Van Luong\inst{1}}
\authorrunning{G.B. Nguyen \textit{et al.}}
%

\institute{Business AI Lab, College of Technology, National Economics University, Vietnam\\
\and
Hanoi University of Science and Technology, Hanoi, Vietnam\\
\and
A2I Lab, Phenikaa School of Computing, Phenikaa University, Hanoi, Vietnam\\
\and
Department of Head and Neck Surgery, Vietnam National Cancer Hospital \& Hanoi Medical University, Hanoi, Vietnam\\
\and
VNPT AI, VNPT Group, Hanoi, Vietnam\\
\email{\{bachng.bai, 11247162\}@st.neu.edu.vn, \{cuong.vuongtuan, trang.maixuan\}@phenikaa-uni.edu.vn, duyyhn@gmail.com, nguyentiencuong@vnpt.vn, \{huanv, thienlv\}@neu.edu.vn}
}

\maketitle              
\begin{abstract}

Clinical predictive models often rely on structured Electronic Health Record data, such as time-series and procedure codes. While recent approaches have begun leveraging unstructured clinical notes, they typically encode them as flat sequences, which may lose explicit relational and temporal structure present in clinical narratives. In response, we propose HERMES, a graph-based framework that operates exclusively on clinical text while preserving clinical relationships. This approach builds on two key ideas. First, personalized Knowledge Graphs (KGs) are constructed through Large-Language-Model-guided extraction from clinical notes with Contrastive Logic Modeling that explicitly captures temporal dynamics and treatment failures and changes in outcomes. Second, a Graph Attention Network synthesizes patient representations through graph-based learning over the KGs. Experiments on MIMIC-III and MIMIC-IV for in-hospital mortality and 30-day readmission prediction show that HERMES consistently outperforms strong text-only baselines. Our findings demonstrate that explicit relational modeling with Contrastive Logic Modeling significantly advances predictive performance.

\end{abstract}
\section{Introduction}
Electronic Health Records (EHRs) enable data-driven risk prediction, including in-hospital mortality and 30-day readmission forecasting. In the ICU, clinical evidence is documented in narrative notes capturing clinician reasoning, patient status, and treatment responses. However, many predictive pipelines prioritize structured signals and treat notes as auxiliary inputs, underutilizing the clinical knowledge embedded in narratives. Recent clinical language models (LMs) \cite{clinicalbert,gu2021pubmedbert} advance outcome prediction from notes, yet most approaches remain token-centric and implicitly represent clinical dependencies through sequence representations. Early prediction models treated clinical narratives as flat sequences, using Convolutional Neural Networks (CNNs), Recurrent Neural Networks (RNNs), and Transformers to represent token streams \cite{clinicalbert,chang2022clinical}. Although long-context variants better accommodate lengthy ICU documentation \cite{chang2022clinical}, these approaches implicitly model clinical dependencies. Hierarchical encoders \cite{li2020notehcr} and token-level graph models introduce local connectivity but remain token-focused. This paradigm cannot explicitly capture clinically meaningful relations, such as symptom-to-diagnosis evidence, treatment-to-response dynamics, and clinical contradictions. Consequently, models fail to expose the structured clinical logic central to adverse outcomes.

Knowledge-enhanced methods advance representations by injecting external medical knowledge from ontologies or curated bases \cite{choi2017gram,jiang2024graphcare}, yet they depend on structured codes rather than free-text. Traditional machine learning approaches, such as logistic regression with TF-IDF \cite{Mahbub}, remain competitive baselines due to simplicity and robustness, but treat text as bags-of-words and cannot capture temporal dependencies. Recent advances in Large Language Models (LLMs) have enabled strong performance in clinical information extraction \cite{agrawal2022large}. Frameworks like EMERGE \cite{zhu2024emerge} and GraphCare \cite{jiang2024graphcare} use LLMs to construct Knowledge Graphs (KGs) for prediction but depend on structured EHR data or external KGs.

In response, this work proposes HERMES, a text-only clinical prediction approach that combines narrative understanding with graph-based reasoning. HERMES differs in three key ways. First, it constructs personalized KGs exclusively from clinical text using LLM-guided extraction with lightweight Named Entity Recognition (NER) model, without external codes or knowledge bases. Second, it employs Contrastive Logic Modeling to prioritize clinical dynamics-temporal changes, treatment contradictions, and adverse outcomes that predict deterioration. Third, it achieves consistent gains over text-only baselines on two ICU benchmarks.

Specifically, HERMES constructs personalized KGs from clinical text through a two-stage extraction pipeline with Contrastive Logic Modeling that emphasizes temporal dynamics. The pipeline extracts entities and their clinical relations while identifying critical patterns-contradictions between findings and treatments, treatment failures, and temporal changes-predictive of adverse outcomes. The KGs are then processed by a Graph Attention Network (GAT) \cite{velickovic2018graph}, producing patient-level representations for outcome prediction.

We evaluate HERMES on MIMIC-III and MIMIC-IV for in-hospital mortality and 30-day readmission prediction. Across both datasets and tasks, the proposed text-only approach consistently outperforms text-only baselines, indicating that explicit relational modeling and attention-based aggregation complement sequence representations. Ablation results characterize the effects of different Graph Neural Network (GNN) backbones and pooling strategies, supporting GAT with max pooling as the optimal encoder.

\section{Methodology}

We consider each patient's first ICU stay from their first visit. Each stay contains a set of clinical notes $\mathcal{N} = \{n_1, n_2, \dots, n_{|\mathcal{N}|}\}$. Given $\mathcal{N}$, we aim to predict the patient's clinical outcome $\hat{y} \in \{0, 1\}$ for two clinical prediction tasks: in-hospital mortality and 30-day readmission. Throughout the framework, we use the concatenated clinical notes, $\mathcal{N}'$, as input to both NER and LLM-based extraction, leveraging the extended context window to preserve semantic relationships across the patient's clinical documentation.

\begin{figure}[!htbp]
    \centering
    \includegraphics[width=1.0\columnwidth]{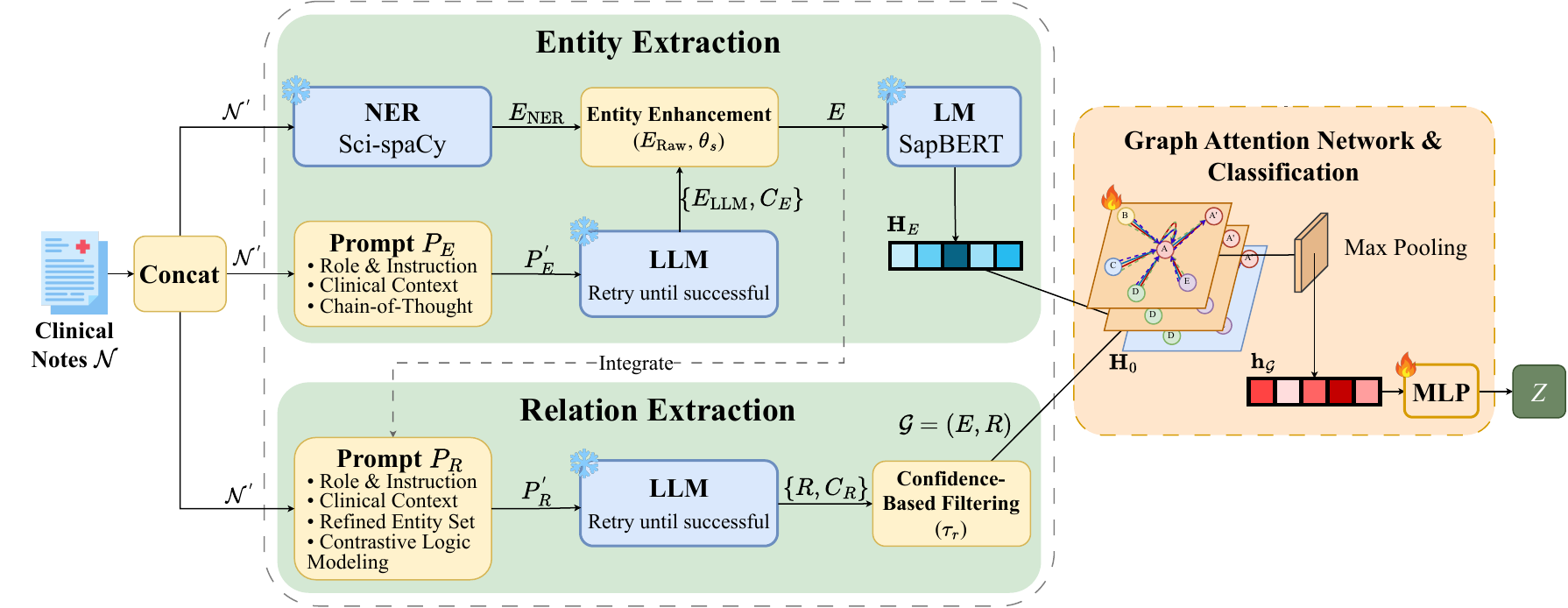}
    \caption{Overall architecture of our proposed HERMES framework. The two extraction modules, Entity Extraction and Relation Extraction, process the input clinical notes $\mathcal{N}$, generating entities $E$ and relations $R$ for each personalized KG. The GAT then synthesizes KG representations for downstream prediction tasks.}
    \label{fig:overview}
\end{figure}

Fig.~\ref{fig:overview} shows the overall architecture of our proposed HERMES framework.

\subsection{Entity Extraction}

Performing NER on unstructured data is challenging. We address this by combining LLM-guided extraction with clinical-domain spaCy NER \footnote{\url{https://allenai.github.io/scispacy/}} to maximize entity coverage. The raw extracted entity set is defined as:

\begin{equation}
E_{\text{raw}} = E_{\text{NER}} \cup E_{\text{LLM}},
\end{equation}
where $E_{\text{NER}}$ and $E_{\text{LLM}}$ represent the entity sets obtained from spaCy-based and LLM-guided extraction, respectively. Each entity is represented as a text string, and similarity comparisons are computed using cosine similarity based on text embeddings.

\subsubsection{Chain-of-Thought Reasoning}

\begin{figure}[!htbp]
    \centering
    \includegraphics[width=1\columnwidth]{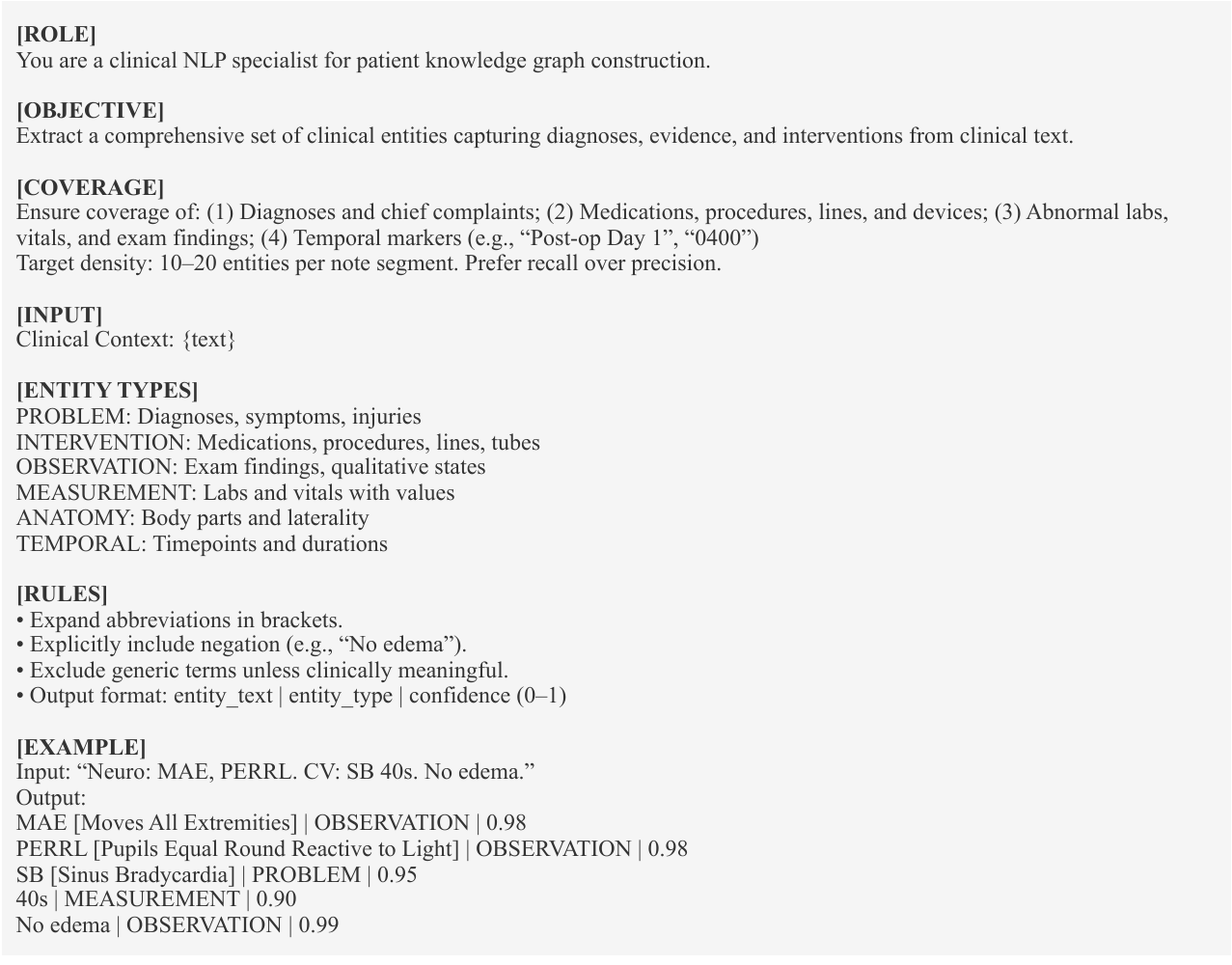}
    \caption{Prompt $P_E$ for entity extraction.}
    \label{fig:entity_prompt}
\end{figure}

Our LLM-guided extraction \cite{agrawal2022large} uses the prompt template $P_E$ (Fig.~\ref{fig:entity_prompt}) that follows a chain-of-thought method \cite{wei2022cot}, requiring the LLM to reason over extractions and assign confidence scores to each entity. Let $C_E$ denote the set of entity confidence scores, where $C_E(e) \in [0, 1]$ represents the confidence score for entity $e$ \cite{tian2023calibration}, which are incorporated into Eq.~\eqref{eq:scoring_ent} and Eq.~\eqref{eq:merge_ent} to guide entity consolidation, prioritizing higher-confidence candidates during merging to form the final entity set $E$. We also preserve negation aspects (for example, ``no pneumothorax'', ``denies pain'') to maintain clinical semantics. Failed extractions are retried until successful, ensuring no entity sets are empty.

\subsubsection{Entity Enhancement}

We normalize the raw entities (whitespace trimming, lowercasing) to unify their representation format. To consolidate the raw entity set $E_{\text{raw}}$ and enhance entity quality, we implement a similarity-based merging algorithm. We define an entity scoring function:

\begin{equation}
\mathrm{Score}(e) = C_E(e) + \mathrm{bonus}(e),
\label{eq:scoring_ent}
\end{equation}
where $C_E(e)$ represents the confidence value assigned by the LLM during extraction (the confidence for NER is fixed), and $\mathrm{bonus}(e)$ denotes an additional bonus term that accounts for extraction source and entity informativeness.

For each pair of entities $e_i, e_j \in E_{\text{raw}}$ where $\mathrm{sim}(e_i, e_j) \ge \theta_s$, we apply:

\begin{equation}
    \texttt{Merge}(e_i, e_j) =
    \begin{cases}
        e_i, & \text{if } \mathrm{Score}(e_i) > \mathrm{Score}(e_j)\\
        e_j, & \text{otherwise}
    \end{cases},
\label{eq:merge_ent}
\end{equation}
where $\theta_s$ represents the similarity threshold. This merging process consolidates similar entities while retaining the higher-scoring candidate according to Eq.~\eqref{eq:merge_ent}, resulting in the final entity set $E$. We then employ a medical-domain language model \cite{liu2021sapbert} to obtain the dense representation for each entity in the set $E$:

\begin{equation}
    \mathbf{H}_E = \mathrm{TextEncoder}(E),
    \label{eq:init_emb}
\end{equation}
where $\mathbf{H}_E$ denotes the entity embeddings of the final set $E$ obtained from the language model.

\subsection{Relation Extraction}

\begin{figure}[!htbp]
    \centering
    \includegraphics[width=1\columnwidth]{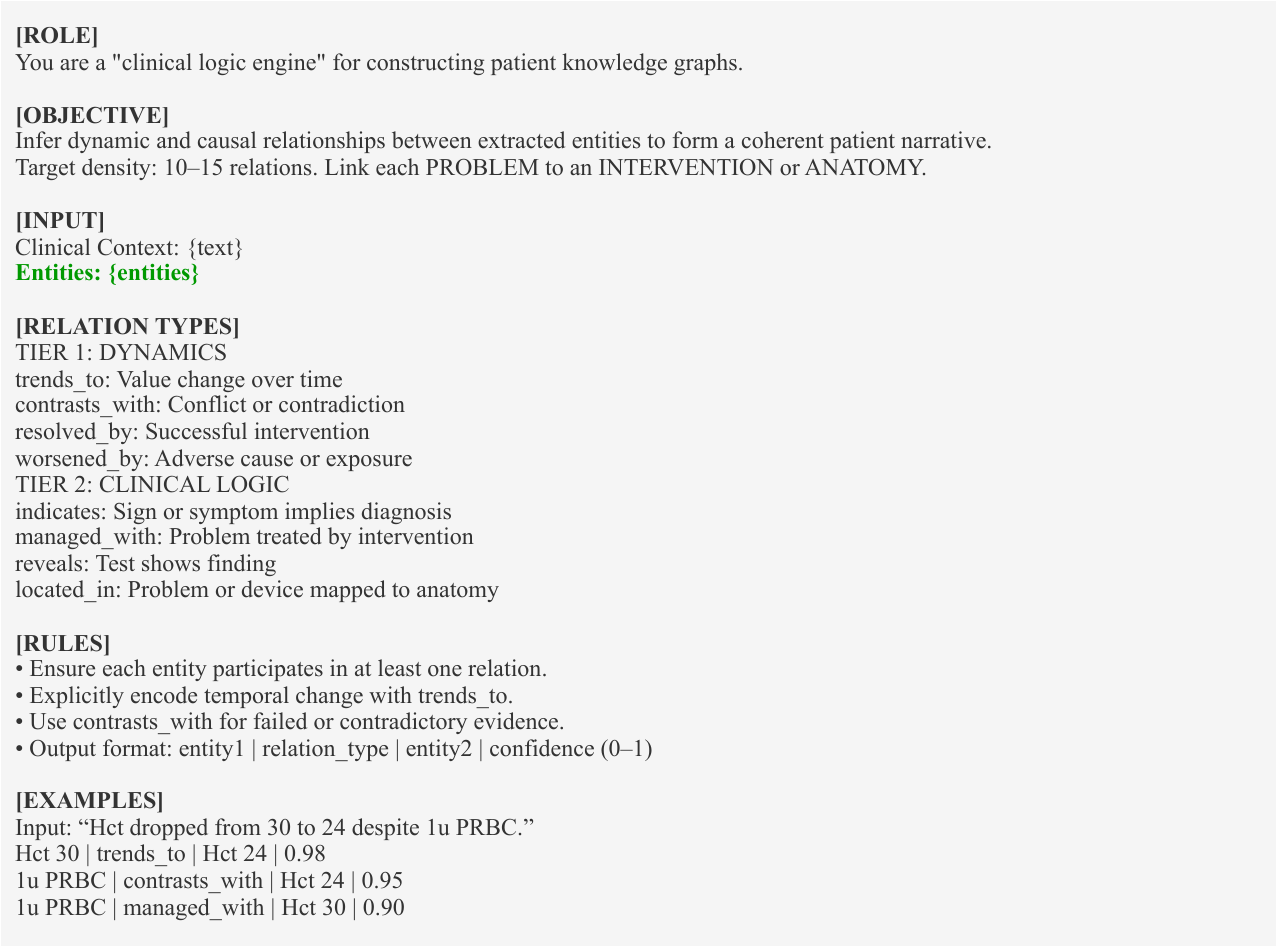}
    \caption{Prompt $P_R$ for relation extraction.}
    \label{fig:relation_prompt}
\end{figure}

While entity extraction identifies key clinical concepts, patient health trajectories depend on how these concepts interact. We employ a second LLM-guided stage for relation extraction, aiming to reconstruct the underlying clinical logic and temporal semantics. We obtain the relation set $R_{\text{LLM}}$ by using the prompt template $P_R$ (Fig.~\ref{fig:relation_prompt}) with $\mathcal{N}'$ and integrating the refined entity set $E$ from the previous stage:

\begin{equation}
    R_{\text{LLM}} = \texttt{LLM}(P_R, \mathcal{N}', E).
\end{equation}

By anchoring relations to extracted entities, we ensure the KG remains grounded in the extracted nodes while allowing the LLM to infer clinical relations from full patient context.

\subsubsection{Contrastive Logic Modeling}

Contrastive Logic Modeling captures clinically important nuances by distinguishing beneficial from harmful treatments. We instruct the LLM to model two relation types: (1) Dynamics capture temporal changes, clinical contradictions, and treatment outcomes-highly predictive of adverse events; (2) Clinical Logic represents static medical knowledge linking observations, diagnoses, interventions, and anatomy. Together, these ensure KGs encode both patient status and temporal evolution.

\begin{figure}[htbp]
    \centering
    \includegraphics[width=0.7\linewidth]{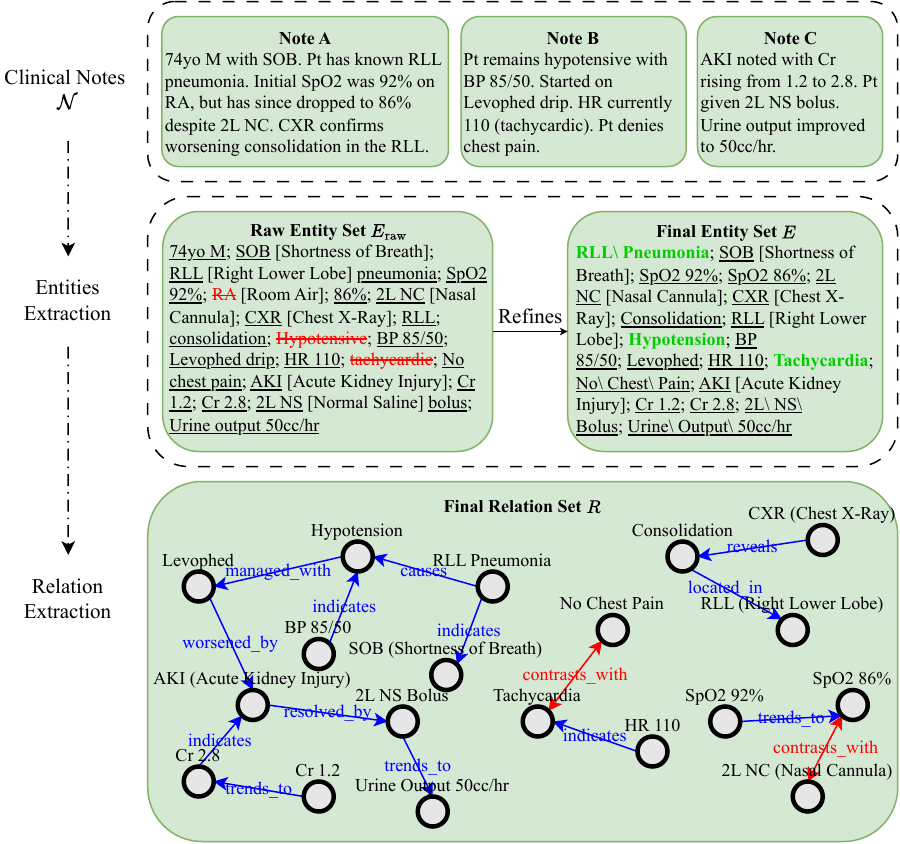}
    \caption{Process of knowledge graph construction in the HERMES framework. \textcolor{blue}{Blue} arrows indicate standard relations, while \textcolor{red}{red} arrows indicate contrastive relations. The contrastive relations are bi-directional by default.}
    \label{fig:example}
\end{figure}

\subsubsection{Confidence-Based Filtering}

To ensure reliable relation extraction for downstream prediction tasks, our relation refinement process employs a confidence-based filtering mechanism \cite{tian2023calibration} to remove ambiguous relations. Let $C_R$ denote the set of relation confidence scores, where $C_R(r) \in [0, 1]$ represents the confidence score for relation $r \in R_{\text{LLM}}$. The final relation set $R$ is defined as:

\begin{equation}
    R = \{ r \in R_{\text{LLM}} \mid C_R(r) \ge \tau_r \},
    \label{eq:relation_set}
\end{equation}
where $R$ denotes the relation set, and $\tau_r$ is the relation confidence threshold. $R$ defines KG connectivity $\mathcal{G} = (E, R)$, transforming unstructured notes into a structured knowledge representation. Contrastive Logic Modeling shapes the KG topology, capturing clinical dependencies and temporal evolution. 

The process of knowledge graph construction is illustrated in Fig.~\ref{fig:example}.

\subsection{Representation Learning}

We encode each patient KG using a two-layer GAT with residual connections and layer normalization, followed by graph pooling and an MLP classifier for outcome prediction. GAT learns attention weights over neighboring nodes, focusing on informative entities without explicitly using relation types.

\subsubsection{Graph Attention Network.}
Let $n$ be the number of entity nodes and $d$ be the node embedding dimension. For $\ell \in \{0,1,2\}$, we define $\mathbf{H}_{\ell} \in \mathbb{R}^{n\times d}$ as the node feature matrix at GAT layer $\ell$, whose $i$-th row is the node embedding $\mathbf{h}_i^{(\ell)} \in \mathbb{R}^{d}$. The input matrix $\mathbf{H}_0 \in \mathbb{R}^{n\times d}$ is initialized by incorporating the entity embeddings $\mathbf{H}_E$ (Eq.~\eqref{eq:init_emb}) with the KG structure $\mathcal{G}$, where each node's initial representation is a vector from $\mathbf{H}_0$. For each node $i$, we define its neighbor set $\mathcal{S}(i)$ as the set of nodes connected to $i$ in the KG, and we include a self-loop so that a node can retain its own information during message passing.

For each layer $\ell \in \{1,2\}$, node $i$, and a neighbor $j\in\mathcal{S}(i)$, GAT first computes an unnormalized attention score $s_{ij}^{(\ell)} \in \mathbb{R}$:
\begin{equation}
    s_{ij}^{(\ell)} = \mathrm{LeakyReLU}\left(\mathbf{a}^T (\mathbf{W}\mathbf{h}_i^{(\ell-1)} \| \mathbf{W}\mathbf{h}_j^{(\ell-1)})\right),
    \label{eq:gat_score}
\end{equation}
where $\mathbf{W} \in \mathbb{R}^{d\times d}$ is a learnable weight matrix, $\mathbf{a} \in \mathbb{R}^{2d}$ is a learnable attention vector, and $\|$ denotes concatenation.
These scores are normalized over the neighborhood of $i$ using a softmax to obtain attention coefficients $\alpha_{ij}^{(\ell)} \in \mathbb{R}$:
\begin{equation}
    \alpha_{ij}^{(\ell)} = \frac{\exp(s_{ij}^{(\ell)})}{\sum_{k \in \mathcal{S}(i)} \exp(s_{ik}^{(\ell)})},
    \label{eq:gat_alpha}
\end{equation}
where the denominator sums over all neighbors $k$ of $i$ (including $i$ itself). The coefficient $\alpha_{ij}^{(\ell)}$ controls how much information node $i$ receives from node $j$.

Node representations are then updated by aggregating transformed neighbor features weighted by the attention coefficients:
\begin{equation}
    \mathbf{h}_i^{(\ell)} = \phi\left(\sum_{j \in \mathcal{S}(i)} \alpha_{ij}^{(\ell)} \mathbf{W} \mathbf{h}_j^{(\ell-1)}\right),
    \label{eq:gat_update}
\end{equation}
where $\phi$ is a non-linear activation function, which is ReLU in our implementation. The framework uses multi-head attention with $K$ heads and concatenates head outputs to form the layer output.

Let $\mathrm{GAT}(\cdot)$ denote applying Eq.~\eqref{eq:gat_update} to all nodes of the layer in parallel. We stack two GAT layers with residual connections and layer normalization:
\begin{equation}
    \mathbf{H}_{\ell} =
    \mathrm{LayerNorm}\!\left(
        \mathbf{H}_{\ell - 1} +
        \mathrm{GAT}\!\left(\mathbf{H}_{\ell - 1}\right)
    \right),
    \label{eq:gat_stack}
\end{equation}
where the initial embeddings are given by $\mathbf{H}_0$. After two layers, $\mathbf{H}_2$ contains the final node representations.

\subsubsection{Pooling and Classification.}
To obtain the graph representation for the patient, we aggregate the node representations in $\mathbf{H}_2$ using max pooling:
\begin{equation}
\mathbf{h}_{\mathcal{G}} =
\mathrm{MaxPool}\!\left(\mathbf{H}_2\right),
\end{equation}
where $\mathbf{h}_{\mathcal{G}} \in \mathbb{R}^{d}$ is the graph representation for the patient. This representation is passed to a two-layer MLP classifier to produce the predicted probability $\hat{y}$:
\begin{equation}
\hat{y} = \sigma\!\left(\mathrm{MLP}(\mathbf{h}_{\mathcal{G}})\right).
\end{equation}
Here, $\mathrm{MLP}(\cdot)$ denotes a two-layer multilayer perceptron that outputs a scalar logit, and $\sigma(\cdot)$ is the sigmoid function for binary classification.

We optimize the model using binary cross-entropy (BCE) loss for both mortality and readmission prediction:
\begin{equation}
\mathcal{L}(\hat{y}, y) = -\frac{1}{M}\sum_{i=1}^{M} \left[y_i \log \hat{y}_i + (1-y_i) \log(1-\hat{y}_i)\right],
\end{equation}
where $M$ is the number of patients, $y_i\in\{0,1\}$ is the ground-truth label, and $\hat{y}_i$ is the predicted probability.

By learning over personalized KGs constructed from clinical notes, the GAT encoder integrates entity information with graph-structured clinical
associations, producing patient representations that support accurate outcome prediction.

\section{Experiments}

\subsection{Experimental Setups}

\subsubsection{Evaluation Datasets}

We evaluate on MIMIC-III \cite{Johnson2016MIMICIII} and MIMIC-IV \cite{Johnson2023MIMICIV} for in-hospital mortality and 30-day readmission prediction. Following the experimental setup of EMERGE \cite{zhu2024emerge}, we use the first 48 hours of each ICU stay, concatenating notes from consecutive 12-hour segments. This setup enables controlled comparison and leverages the extended context window to capture clinical trajectories. The datasets statistics are shown in Table~\ref{tab:dataset_stats}. $\textbf{Label}_{\text{Mortality}}$ and $\textbf{Label}_{\text{Readmission}}$ denote the number (and percentage) of positive labels for in-hospital mortality and 30-day readmission, respectively.

\begin{table}[!htbp]
\centering
\caption{Dataset statistics after preprocessing}
\label{tab:dataset_stats}

\setlength{\tabcolsep}{7pt}
\renewcommand{\arraystretch}{1.15}

\resizebox{0.65\linewidth}{!}{%
\begin{tabular}{llccc}
\toprule
\textbf{Dataset} & \textbf{Split} & \textbf{Samples} & $\textbf{Label}_{\text{Mortality}}$ & $\textbf{Label}_{\text{Readmission}}$ \\
\midrule
\multirow{3}{*}{MIMIC-III}
& Train & 11{,}200 (70.00\%) & 1{,}538 (13.7\%) & 2{,}024 (18.1\%) \\
& Val   & 1{,}600  (10.00\%) & 220 (13.8\%)    & 289 (18.1\%) \\
& Test  & 3{,}200  (20.00\%) & 440 (13.8\%)    & 578 (18.1\%) \\
\midrule
\multirow{3}{*}{MIMIC-IV}
& Train & 11{,}200 (70.00\%) & 2{,}082 (18.6\%) & 2{,}741 (24.5\%) \\
& Val   & 1{,}600  (10.00\%) & 297 (18.6\%)    & 391 (24.4\%) \\
& Test  & 3{,}200  (20.00\%) & 595 (18.6\%)    & 783 (24.5\%) \\
\bottomrule
\end{tabular}%
}
\end{table}

The two datasets differ substantially in documentation density: MIMIC-III contains 168,194 total notes (mean 10.5 per patient) with 221.5 words per note on average, while MIMIC-IV has 51,297 notes (mean 3.2 per patient) with 196.4 words per note. This translates to significantly denser extracted KGs, where MIMIC-III yields a mean of 133.7 entities and 1,263.4 relations per patient compared to 47.4 entities and 61.7 relations in MIMIC-IV. This reflects distributional differences between the two datasets.

\subsubsection{Evaluation Metrics}

We evaluate using AUROC (threshold-independent discrimination), AUPRC (emphasizes positive-class performance under imbalance), and min(+P, Se) (minimum of precision and sensitivity). Higher values indicate better performance. In all results tables, \textbf{Bold} and \underline{Underlined} denote best and second-best results, respectively.

\subsubsection{Baseline Models}

We compare our approach against a diverse set of baseline methods. These include pre-trained clinical language models such as ClinicalBERT \cite{clinicalbert}, PubMedBERT \cite{gu2021pubmedbert}, and Clinical-Longformer \cite{chang2022clinical}; Note-HCR \cite{li2020notehcr}, a hierarchical RNN-based model; and Logistic Regression with TF-IDF features \cite{Mahbub}, a traditional machine learning baseline that can process full documents but lacks semantic understanding. We additionally include a zero-shot LLM prompting baseline (GPT-5.4 mini\footnote{\url{https://developers.openai.com/api/docs/models/gpt-5.4-mini}}) to assess the ceiling of general-purpose LLM reasoning on this task. We further evaluate different GNN variants, including Graph Convolutional Network (GCN), GAT, and Relational Graph Convolutional Network (RGCN), to assess their effectiveness.

\subsubsection{Implementation Details}

Experiments are conducted on a single NVIDIA RTX 3060 GPU (12\,GB). We use SapBERT \cite{liu2021sapbert} for entity embedding and Mistral-Small-3.2-24B-Instruct\footnote{\url{https://docs.mistral.ai/models/mistral-small-3-2-25-06}} for LLM-guided extraction \cite{agrawal2022large}, both models are kept frozen. The GAT encoder has two layers with eight attention heads and ReLU activation, hidden dimension 128, and max pooling. The two-layer MLP classifier uses GELU \cite{hendrycks2016gelu} activation. Training uses AdamW \cite{loshchilov2019adamw} with learning rate 0.001, batch size 64, dropout 0.2, and patience 10. For extraction thresholds, we set entity similarity $\theta_s = 0.85$ and relation confidence $\tau_r = 0.7$. Results are reported as $\text{mean}\pm\text{std}$ via bootstrapping (1{,}000 resamples).

\subsection{Experimental Results}

We compare HERMES against representative text-only baselines on in-hospital mortality and 30-day readmission prediction in MIMIC-III and MIMIC-IV, as shown in Table~\ref{tab:baselines}.

\begin{table}[!htbp]
\centering
\caption{Performance comparison between HERMES and different baselines on MIMIC-III and MIMIC-IV datasets}
\label{tab:baselines}

\setlength{\tabcolsep}{4.5pt}
\renewcommand{\arraystretch}{1.1}

\resizebox{0.9\linewidth}{!}{
\begin{tabular}{ll|ccc|ccc}
\toprule
\multirow{2}{*}{\textbf{Dataset}} & \multirow{2}{*}{\textbf{Method}} &
\multicolumn{3}{c|}{\textbf{Mortality}} &
\multicolumn{3}{c}{\textbf{Readmission}} \\
& & AUROC($\uparrow$) & AUPRC($\uparrow$) & min(+P, Se)($\uparrow$) &
  AUROC($\uparrow$) & AUPRC($\uparrow$) & min(+P, Se)($\uparrow$) \\
\midrule
\multirow{6}{*}{MIMIC-III}
& Clinical-LongFormer \cite{chang2022clinical} & 70.93$\pm$0.96 & 25.84$\pm$1.58 & 28.62$\pm$1.42 & 66.60$\pm$0.97 & 28.89$\pm$1.55 & 31.27$\pm$1.60 \\
& ClinicalBERT \cite{clinicalbert}       & 70.99$\pm$2.44 & 26.95$\pm$3.77 & 29.58$\pm$3.86 & 67.72$\pm$2.35 & 30.50$\pm$3.43 & 33.04$\pm$3.26 \\
& PubMedBERT \cite{gu2021pubmedbert}         & 67.43$\pm$2.74 & 23.72$\pm$3.09 & 27.95$\pm$3.65 & 64.37$\pm$2.49 & 26.88$\pm$2.89 & 30.69$\pm$3.47 \\

& GPT-5.4 mini (Zero-shot)    & 72.36$\pm$1.17 & 32.83$\pm$1.90 & 49.47$\pm$2.25 & 50.10$\pm$0.19 & 18.10$\pm$0.68 & 18.06$\pm$0.67 \\

& Note-HCR \cite{li2020notehcr}           & 81.99$\pm$1.35 & 49.63$\pm$3.00 & 47.86$\pm$2.00 & 74.41$\pm$1.20 & 45.68$\pm$1.83 & 44.78$\pm$1.86 \\
& LR + TF-IDF \cite{Mahbub}        & \underline{83.54$\pm$1.02} & \underline{50.67$\pm$2.55} & \underline{48.63$\pm$2.14} & \underline{76.76$\pm$1.12} & \underline{47.83$\pm$2.19} & \underline{47.10$\pm$1.85} \\
& HERMES (Ours)              & \textbf{85.92$\pm$0.94} & \textbf{55.49$\pm$2.53} & \textbf{51.52$\pm$2.12} & \textbf{77.75$\pm$1.11} & \textbf{49.92$\pm$2.16} & \textbf{47.25$\pm$1.88} \\
\midrule
\multirow{6}{*}{MIMIC-IV}
& Clinical-LongFormer \cite{chang2022clinical} & 69.51$\pm$0.94 & 29.45$\pm$1.42 & 33.58$\pm$2.22 & 65.89$\pm$1.03 & 35.30$\pm$1.48 & 37.48$\pm$1.42 \\
& ClinicalBERT \cite{clinicalbert}       & 75.51$\pm$2.01 & 43.95$\pm$4.01 & 43.47$\pm$3.47 & 69.57$\pm$2.11 & 44.60$\pm$3.49 & 44.52$\pm$3.12 \\
& PubMedBERT \cite{gu2021pubmedbert}         & 74.05$\pm$2.14 & 43.37$\pm$4.07 & 44.02$\pm$3.43 & 70.13$\pm$2.09 & 44.70$\pm$3.71 & 44.20$\pm$3.27 \\

& GPT-5.4 mini (Zero-shot)    & 71.32$\pm$0.98 & 31.07$\pm$1.30 & 35.94$\pm$1.39 & 50.45$\pm$0.30 & 24.70$\pm$0.81 & 24.47$\pm$0.78 \\

& Note-HCR \cite{li2020notehcr}           & 80.47$\pm$0.91 & 51.78$\pm$2.02 & 49.07$\pm$1.72 & 73.20$\pm$1.03 & 48.94$\pm$1.85 & 47.50$\pm$1.66 \\
& LR + TF-IDF \cite{Mahbub}        & \underline{81.83$\pm$0.85} & \underline{55.27$\pm$1.98} & \underline{50.56$\pm$1.69} & \underline{74.78$\pm$1.01} & \underline{52.62$\pm$1.81} & \underline{49.54$\pm$1.65} \\
& HERMES (Ours)              & \textbf{83.24$\pm$0.89} & \textbf{58.83$\pm$1.95} & \textbf{53.00$\pm$1.70} & \textbf{77.31$\pm$0.99} & \textbf{56.88$\pm$1.87} & \textbf{53.73$\pm$1.53} \\
\bottomrule
\end{tabular}
}
\end{table}

Overall, HERMES achieves the best results across both datasets and tasks. Compared with the strongest non-graph baseline \cite{Mahbub}, HERMES improves AUROC/AUPRC by +2.38/+4.82 points and +1.41/+3.56 points on MIMIC-III and MIMIC-IV mortality, respectively, while also yielding consistent gains for readmission (+2.53 AUROC and +4.26 AUPRC on MIMIC-IV). These improvements suggest that explicitly modeling patient-specific relations extracted from notes provides a strong predictive signal.

It is also worth noting that the traditional ML baseline \cite{Mahbub} remains quite competitive, likely due to its ability to capture key prognostic cues from early ICU notes. However, unlike HERMES, this performance is driven primarily by surface-level lexical signals rather than the underlying clinical and temporal relationships, such as evolving findings, treatment failures, and contradictions. This is further evidenced by the zero-shot prompting baseline. Despite superior reasoning capability and broad medical knowledge, it achieves only 72.36 AUROC on MIMIC-III mortality, marginally above the pre-trained clinical language models, yet substantially below simple ML baseline\cite{Mahbub}. This confirms that zero-shot reasoning, regardless of model scale, cannot replace structured, task-specific learning over explicit clinical relations. Bridging this gap via prompting alone would likely require frontier LLM models with more complex reasoning pipelines, which are expensive and impractical for clinical deployment.

\subsection{Ablation Studies}

\subsubsection{Comparing Different GNN Backbones}

We study the impact of the graph encoder by swapping the GNN backbone and pooling strategy while keeping the framework fixed. We compare GCN, GAT, and RGCN under mean and max pooling, evaluated on the same prediction tasks, as shown in Table~\ref{tab:gnn_ablation}.

\begin{table}[!htbp]
\centering
\caption{Performance comparison of different GNNs in HERMES on MIMIC-III and MIMIC-IV datasets}
\label{tab:gnn_ablation}

\setlength{\tabcolsep}{4.5pt}
\renewcommand{\arraystretch}{1.1}

\resizebox{0.9\linewidth}{!}{
\begin{tabular}{ll|ccc|ccc}
\toprule
\multirow{2}{*}{\textbf{Dataset}} & \multirow{2}{*}{\textbf{Method}} &
\multicolumn{3}{c|}{\textbf{Mortality}} &
\multicolumn{3}{c}{\textbf{Readmission}} \\
& & AUROC($\uparrow$) & AUPRC($\uparrow$) & min(+P, Se)($\uparrow$) &
  AUROC($\uparrow$) & AUPRC($\uparrow$) & min(+P, Se)($\uparrow$) \\
\midrule
\multirow{6}{*}{MIMIC-III}
& GCN (mean) & 82.31$\pm$1.02 & 45.88$\pm$2.57 & 44.74$\pm$2.16 & 76.50$\pm$1.08 & 45.23$\pm$2.26 & 44.63$\pm$1.88 \\
& GCN (max)  & 83.61$\pm$1.05 & \underline{52.02$\pm$2.59} & \underline{49.16$\pm$2.12} & \underline{77.61$\pm$1.09} & \underline{49.84$\pm$2.17} & \textbf{47.41$\pm$1.87} \\
& GAT (mean) & 83.04$\pm$1.06 & 49.65$\pm$2.63 & 48.26$\pm$2.12 & 76.87$\pm$1.12 & 46.37$\pm$2.23 & 45.83$\pm$1.90 \\
& GAT (max)  & \textbf{85.92$\pm$0.94} & \textbf{55.49$\pm$2.53} & \textbf{51.52$\pm$2.12} & \textbf{77.75$\pm$1.11} & \textbf{49.92$\pm$2.16} & \underline{47.25$\pm$1.88} \\
& RGCN (mean)& 83.09$\pm$1.03 & 47.69$\pm$2.60 & 48.01$\pm$2.05 & 76.39$\pm$1.10 & 46.52$\pm$2.22 & 45.04$\pm$1.84 \\
& RGCN (max) & \underline{83.14$\pm$1.04} & 49.52$\pm$2.44 & 46.48$\pm$2.15 & 76.44$\pm$1.12 & 48.08$\pm$2.14 & 46.43$\pm$1.88 \\
\midrule
\multirow{6}{*}{MIMIC-IV}
& GCN (mean) & 80.26$\pm$0.93 & 51.13$\pm$2.12 & 49.42$\pm$1.78 & 74.29$\pm$1.07 & 51.71$\pm$1.94 & 48.81$\pm$1.67 \\
& GCN (max)  & \underline{82.65$\pm$0.86} & \underline{57.30$\pm$1.91} & \underline{52.35$\pm$1.71} & 76.07$\pm$1.00 & \underline{54.83$\pm$1.83} & 51.38$\pm$1.57 \\
& GAT (mean) & 80.71$\pm$0.93 & 51.67$\pm$2.12 & 48.49$\pm$1.84 & 74.04$\pm$1.05 & 49.27$\pm$1.95 & 47.76$\pm$1.68 \\
& GAT (max)  & \textbf{83.24$\pm$0.89} & \textbf{58.83$\pm$1.95} & \textbf{53.00$\pm$1.70} & \textbf{77.31$\pm$0.99} & \textbf{56.88$\pm$1.87} & \textbf{53.73$\pm$1.53} \\
& RGCN (mean)& 80.06$\pm$0.91 & 49.29$\pm$2.12 & 46.60$\pm$1.79 & 72.27$\pm$1.03 & 45.86$\pm$1.93 & 45.04$\pm$1.73 \\
& RGCN (max) & 81.70$\pm$0.95 & 55.73$\pm$2.03 & 50.72$\pm$1.80 & \underline{76.36$\pm$1.02} & 54.23$\pm$1.91 & \underline{51.39$\pm$1.64} \\
\bottomrule
\end{tabular}
}
\end{table}
GAT with max pooling consistently performs best across both datasets and tasks, indicating that attention-based neighborhood aggregation is effective for highlighting clinically informative entities and relations. Across backbones, max pooling generally outperforms mean pooling, suggesting that a small set of high-signal nodes often drives the prediction.

\section{Conclusion}

This work proposes HERMES, a novel framework that constructs personalized KGs from clinical text through LLM-guided extraction with Contrastive Logic Modeling, which captures temporal dynamics, such as treatment contradictions, clinical outcomes, and patient changes-that predict deterioration. Experiments on MIMIC-III and MIMIC-IV demonstrate significant gains over text-only baselines for mortality and readmission prediction, validating explicit relational modeling for clinical outcome prediction. Future work may integrate multimodal data, refine the extraction pipeline, and validate on external hospital datasets to assess generalizability. Intrinsic KG quality evaluation also remains a worthwhile direction for further validation.

\section*{Acknowledgement}
This research is supported by the Ben Dam Me Award Fund, the Vietnam Young Talent Support Fund, and the Number One Brand, Tan Hiep Phat Group.
%
\bibliographystyle{splncs04-modified}
\bibliography{references}

@String{Academic = "Academic Press" }

@String{Chelsea = "Chelsea" }

@String{ICLR = "Proc. Int. Conf. Learn. Represent."}

@String{EMNLP = "Proc. Conf. Empirical Methods Nat. Lang. Process."}

@String{NAACL = "Proc. Conf. North Amer. Chapter Assoc. Comput. Linguist."}

@String{KDD = "Proc. ACM SIGKDD Int. Conf. Knowl. Discov. Data Min."}

@String{CIKM = "Proc. ACM Int. Conf. Inf. Knowl. Manag."}

@article{clinicalbert,
  author  = {Wang, G. and Liu, X. and Liu, H. and Yang, G. and others},
  title   = {A Generalist Medical Language Model for Disease Diagnosis Assistance},
  journal = {Nature Medicine},
  year    = {2025}
}

@article{Johnson2016MIMICIII,
  title={MIMIC-III, a freely accessible critical care database},
  author={Johnson, Alistair EW and Pollard, Tom J and Shen, Lu and Lehman, Li-wei H and Feng, Mengling and Ghassemi, Mohammad and Moody, Benjamin and Szolovits, Peter and Anthony Celi, Leo and Mark, Roger G},
  journal={Scientific data},
  volume={3},
  number={1},
  pages={1--9},
  year={2016},
  publisher={Nature Publishing Group}
}

@article{Johnson2023MIMICIV,
  title={MIMIC-IV, a freely accessible electronic health record dataset},
  author={Johnson, Alistair EW and Bulgarelli, Lucas and Shen, Lu and Gayles, Alvin and Shammout, Ayad and Horng, Steven and Pollard, Tom J and Hao, Sicheng and Moody, Benjamin and Gow, Brian and others},
  journal={Scientific data},
  volume={10},
  number={1},
  pages={1},
  year={2023},
  publisher={Nature Publishing Group UK London}
}

@inproceedings{velickovic2018graph,
    title={Graph Attention Networks},
    author={Petar Veličković and Guillem Cucurull and Arantxa Casanova and Adriana Romero and Pietro Liò and Yoshua Bengio},
    booktitle=ICLR,
    year={2018},
}

@inproceedings{liu2021sapbert,
  title={Self-alignment pretraining for biomedical entity representations},
  author={Liu, Fangyu and Shareghi, Ehsan and Meng, Zaiqiao and Basaldella, Marco and Collier, Nigel},
  booktitle=NAACL,
  pages={4228--4238},
  year={2021}
}

@article{wei2022cot,
  title={Chain-of-thought prompting elicits reasoning in large language models},
  author={Wei, Jason and Wang, Xuezhi and Schuurmans, Dale and Bosma, Maarten and Xia, Fei and Chi, Ed and Le, Quoc V and Zhou, Denny and others},
  journal={Advances in neural information processing systems},
  volume={35},
  pages={24824--24837},
  year={2022}
}

@inproceedings{tian2023calibration,
    title = {Just Ask for Calibration: Strategies for Eliciting Calibrated Confidence Scores from Language Models Fine-Tuned with Human Feedback},
    author = {Tian, Katherine  and Mitchell, Eric  and Zhou, Allan  and Sharma, Archit  and Rafailov, Rafael  and Yao, Huaxiu  and Finn, Chelsea  and Manning, Christopher},
    booktitle = EMNLP,
    year = {2023}
}

@article{gu2021pubmedbert,
    author = {Gu, Yu and Tinn, Robert and Cheng, Hao and Lucas, Michael and Usuyama, Naoto and Liu, Xiaodong and Naumann, Tristan and Gao, Jianfeng and Poon, Hoifung},
    title = {Domain-Specific Language Model Pretraining for Biomedical Natural Language Processing},
    year = {2021},
    volume = {3},
    number = {1},
    journal = {ACM Trans. Comput. Healthcare},
}

@article{hendrycks2016gelu,
  title={Gaussian Error Linear Units (GELUs)},
  author={Dan Hendrycks and Kevin Gimpel},
  journal={arXiv arXiv:1606.08415
        
        
        
        
        
        },
  year={2016},
}

@article{chang2022clinical,
  title={A comparative study of pretrained language models for long clinical text},
  author={Li, Yikuan and Wehbe, Ramsey M and Ahmad, Faraz S and Wang, Hanyin and Luo, Yuan},
  journal={Journal of the American Medical Informatics Association},
  volume={30},
  number={2},
  pages={340--347},
  year={2023},
  publisher={Oxford Academic}
}

@inproceedings{agrawal2022large,
    title = {Large language models are few-shot clinical information extractors},
    author = {Agrawal, Monica  and Hegselmann, Stefan  and Lang, Hunter  and Kim, Yoon  and Sontag, David},
    booktitle = EMNLP,
    year = {2022},
    pages = {1998--2022}
}

@inproceedings{loshchilov2019adamw,
  title     = {Decoupled Weight Decay Regularization},
  author    = {Ilya Loshchilov and Frank Hutter},
  booktitle = ICLR,
  year      = {2019},
}

@inproceedings{zhu2024emerge,
  title={Emerge: Enhancing multimodal electronic health records predictive modeling with retrieval-augmented generation},
  author={Zhu, Yinghao and Ren, Changyu and Wang, Zixiang and Zheng, Xiaochen and Xie, Shiyun and Feng, Junlan and Zhu, Xi and Li, Zhoujun and Ma, Liantao and Pan, Chengwei},
  booktitle=CIKM,
  pages={3549--3559},
  year={2024}
}

@inproceedings{jiang2024graphcare,
    title={GraphCare: Enhancing Healthcare Predictions with Personalized Knowledge Graphs},
    author={Pengcheng Jiang and Cao Xiao and Adam Richard Cross and Jimeng Sun},
    booktitle=ICLR,
    year={2024}
}

@inproceedings{choi2017gram,
  title={GRAM: graph-based attention model for healthcare representation learning},
  author={Choi, Edward and Bahadori, Mohammad Taha and Song, Le and Stewart, Walter F and Sun, Jimeng},
  booktitle=KDD,
  pages={787--795},
  year={2017}
}

@article{li2020notehcr,
title = {Towards unstructured mortality prediction with free-text clinical notes},
journal = {Journal of Biomedical Informatics},
volume = {108},
pages = {103489},
year = {2020},
author = {Mohammad Hashir and Rapinder Sawhney}
}

@article{Mahbub,
    author = {Mahbub, Maria and Srinivasan, Sudarshan and Danciu, Ioana and Peluso, Alina and Begoli, Edmon and Tamang, Suzanne and Peterson, Gregory},
    year = {2022},
    pages = {e0262182},
    title = {Unstructured clinical notes within the 24 hours since admission predict short, mid \& long-term mortality in adult ICU patients},
    volume = {17},
    journal = {PLOS ONE}
}

\end{document}